\documentclass{article}

\PassOptionsToPackage{numbers,compress,}{natbib}
\usepackage[final]{neurips_2022}

\usepackage[utf8]{inputenc} %
\usepackage[T1]{fontenc}    %
\usepackage[colorlinks,pagebackref]{hyperref}       %
\usepackage{url}            %
\usepackage{booktabs}       %
\usepackage{amsfonts}       %
\usepackage{nicefrac}       %
\usepackage{microtype}      %
\usepackage[dvipsnames]{xcolor}         %

\usepackage{amsmath}
\usepackage{enumitem}
\usepackage[capitalise]{cleveref}
\usepackage{makecell}
\usepackage{graphicx}
\usepackage{lipsum}

\usepackage{colortbl}
\usepackage{subcaption}

\newcolumntype{g}{>{\columncolor{Gray}}c}

\definecolor{Gray}{gray}{0.95}
\definecolor{myblue}{HTML}{74ABDC}
\definecolor{mygreen}{HTML}{86BA64}
\definecolor{myyellow}{HTML}{FFC000}
\definecolor{myorange}{HTML}{ED7D31}

\usepackage{pifont}%
\newcommand{\cmark}{\ding{51}}%
\newcommand{\xmark}{\ding{55}}%

\newcommand{\myquote}[1]{\textit{``#1''}}

\title{ZSON: Zero-Shot Object-Goal Navigation using Multimodal Goal Embeddings}

\author{%
  Arjun Majumdar\thanks{equal contribution},\hspace{0.5em}
  Gunjan Aggarwal\footnotemark[1],\hspace{0.5em}
  Bhavika Devnani,\hspace{0.5em}
  Judy Hoffman,\hspace{0.5em}
      Dhruv Batra\\
  \AND
  \vspace{-4ex}\\
  Georgia Institute of Technology \\
  \vspace{-1ex}\\
  {\small \texttt{\url{https://github.com/gunagg/zson}} } \\
}

\begin{document}

\maketitle

\begin{abstract}
We present a scalable approach for learning \emph{open-world} object-goal navigation (\texttt{ObjectNav}) -- the task of asking a virtual robot (agent) to find any instance of an object in an unexplored environment (e.g., \myquote{find a sink}). Our approach is entirely \emph{zero-shot} -- i.e., it does not require \texttt{ObjectNav} rewards or demonstrations of any kind. Instead, we train on the image-goal navigation (\texttt{ImageNav}) task, in which agents find the location where a picture (i.e., goal image) was captured. Specifically, we encode goal images into a multimodal, semantic embedding space to enable training semantic-goal navigation (\texttt{SemanticNav}) agents at scale in unannotated 3D environments (e.g., HM3D). After training, \texttt{SemanticNav} agents can be instructed to find objects described in free-form natural language (e.g., \myquote{sink,} \myquote{bathroom sink,} etc.) by projecting language goals into the same multimodal, semantic embedding space. As a result, our approach enables open-world \texttt{ObjectNav}. We extensively evaluate our agents on three \texttt{ObjectNav} datasets (Gibson, HM3D, and MP3D) and observe absolute improvements in success of 4.2\% - 20.0\% over existing zero-shot methods. For reference, these gains are similar or better than the 5\% improvement in success between the Habitat 2020 and 2021 \texttt{ObjectNav} challenge winners. In an open-world setting, we discover that our agents can generalize to compound instructions with a room explicitly mentioned (e.g., \myquote{Find a kitchen sink}) and when the target room can be inferred (e.g., \myquote{Find a sink and a stove}).
\end{abstract}

\begin{figure}[ht]
  \centering
  \includegraphics[width=0.9\textwidth]{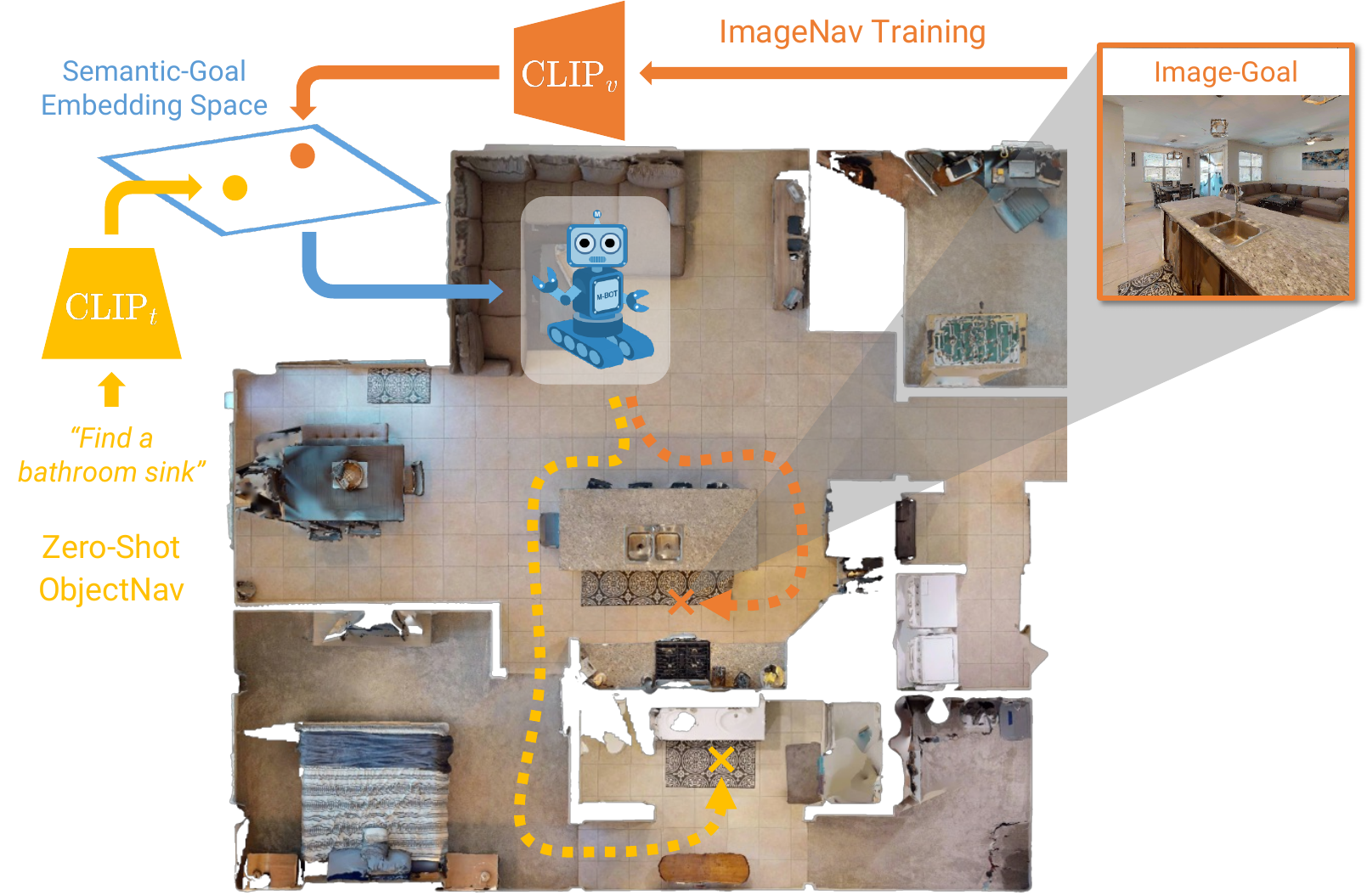}
  \caption{We propose projecting navigation goals (from images or text) into a common, semantic embedding space using a pre-trained vision and language model (CLIP). This allows agents trained with image-goals to understand goals expressed in free-form natural language (e.g., \textcolor{myyellow}{\myquote{Find a bathroom sink.}}). Accordingly, our approach enables \emph{open-world} object-goal navigation in a \emph{zero-shot} manner -- i.e., without using ObjectNav rewards or demonstrations for training.}
  \label{fig:teaser}
\end{figure}

\section{Introduction}

Imagine asking a home assistant robot to find a \myquote{flat-head screwdriver} or the \myquote{medicine case near the bathroom sink.} Building such assistive agents is a problem of deep scientific and societal value.

To study this problem systematically, the embodied AI community has rallied around a problem called object-goal navigation ( \texttt{ObjectNav})~\cite{batra2020objectnav}. Given the name of an object (e.g., \myquote{chair}), \texttt{ObjectNav} involves exploring a 3D environment to find any instance of the object. The last few years have witnessed the development of new environments~\cite{savva2019habitat, szot2021habitat, xia2018gibson, ai2thor, talbot2020benchbot}, annotated 3D scans~\cite{shapenet2015, chang2017mp3d, armeni20193d}, datasets of human demonstrations~\cite{rramrakhya2022habitatweb}, and approaches for ObjectNav~\cite{chaplot2020object, ye2021auxiliary, maksymets2021thda, liang2021sscnav, luo2022stubborn, yadav2022ovrl}, cumulatively leading to strong progress. For instance, the entries in the annual Habitat challenge~\cite{habitatchallenge2022} have jumped from 6\% success (DD-PPO baseline in 2020) to 53\% success (top entry in ongoing 2022 Habitat Challenge public leaderboard).

While this progress is exciting, we believe that a subtle but insidious assumption has snuck into this line of work: the closed-world assumption. We started by discussing an open-world scenario where a person may describe any object in language (e.g., \myquote{flat-head screwdriver}), but \texttt{ObjectNav} is currently formulated over a closed predetermined vocabulary of object categories (\myquote{chair}, \myquote{bed}, \myquote{sofa}, etc.), with approaches using pre-trained object detectors and segmenters for these categories~\cite{rramrakhya2022habitatweb, chaplot2020object, ye2021auxiliary, maksymets2021thda}. While this assumption may have been essential to get started on this problem, it is now important to move beyond it and ask -- how can embodied agents find objects in an open-world setting? 

In this work, we develop an approach for \texttt{ObjectNav} that is both \emph{zero-shot}, i.e., does not require \emph{any} \texttt{ObjectNav} rewards or demonstrations, and \emph{open-world}, i.e., does not require committing to a taxonomy of categories. Our key insight is that we can create a visiolinguistic embedding space to decouple two problems -- (1) describing and representing semantic goals (\myquote{chair}, \myquote{brown chair}, picture of brown chair) from (2) learning to navigate to semantic goals.\footnote{Similar arguments have been made by~\citet{al2022zero}. A detailed discussion is provided in~\cref{sec:related-work}.}

\looseness=-1
To represent semantic goals (1), we leverage recent advances in multimodal AI research on learning a common embedding space for images and text using large collections of image-captions pairs. Specifically, we use CLIP~\cite{radford2021clip}, a method for training dual vision and language encoders that produce similar representations for paired data such as an image and its caption. As shown in~\cref{fig:teaser}, we use CLIP to transform image-goals (e.g., a picture of the kitchen island) and object-goals (e.g., \myquote{bathroom sink}) into \emph{semantic-goals} representing navigation targets. Our main observation is that a semantic-goal produced from an image (e.g., a picture of the bathroom sink) should be similar to semantic goals produced from descriptions of the same target (e.g, \myquote{bathroom sink}). Thus, we hypothesize that these modalities (images and language) can be used interchangeably for creating semantic goals.

Accordingly, for learning to navigate to semantic goals (2), we train agents using image-goals encoded via CLIP's image encoder. Then, we evaluate the learned navigation policy on \texttt{ObjectNav}, where goals are specified in language (e.g., \myquote{chair}) and encoded via CLIP's text encoder. As a result, our agents perform \texttt{ObjectNav} without ever directly training for the task -- i.e., in a zero-shot manner.

\looseness=-1
An important advantage of our approach is that it reduces the data labeling burden. Image-goals can be procedurally generated by randomly sampling points in 3D environments. This is in stark contrast to \texttt{ObjectNav}, which requires annotating 3D meshes~\cite{shapenet2015, chang2017mp3d, armeni20193d} and potentially collecting large-scale human demonstrations~\cite{rramrakhya2022habitatweb} for training. Secondly, the interface to our agents is a natural language description -- matching the grand vision that inspired the \texttt{ObjectNav} task. Through this interface we can refine object-goals by, for instance, specifying object attributes (\myquote{brown chair}) or indicating which room the object is in (\myquote{bathroom sink}) -- which is not possible with traditional \texttt{ObjectNav} agents.

We perform large-scale experiments on three \texttt{ObjectNav} datasets -- Gibson~\cite{xia2018gibson}, MP3D~\cite{chang2017mp3d}, and HM3D~\cite{ramakrishnan2021hm3d}. Our zero-shot agent (that has not seen a single 3D semantic annotation or \texttt{ObjectNav} training episode) achieves a 31.3\% success in Gibson environments, which is a 20.0\% absolute improvement over previous zero-shot results~\cite{al2022zero}. In MP3D, our agent achieves 15.3\% success, a 4.2\% absolute gain over existing zero-shot methods\cite{gadre2022cow}. For reference, these gains are on par or better than the 5\% improvement in success between the Habitat 2020 and 2021 \texttt{ObjectNav} challenge winners. On HM3D, our agent's zero-shot SPL matches a state-of-the-art \texttt{ObjectNav} method~\cite{yadav2022ovrl} that trains with direct supervision from 40k human demonstrations.

Additionally, we study two techniques that are used in our approach to improve zero-shot \texttt{ObjectNav} performance. First, we find that pretraining the visual observation encoder has an outsized effect on zero-shot transfer. Specifically, success on the \texttt{ImageNav} training task improves 4.5\% - 5.8\%, while downstream success on zero-shot \texttt{ObjectNav} improves by 9.4\% - 10.4\%. Similarly, increasing the number of training environments (from 72 to 800) leads to a small drop in \texttt{ImageNav} success, but results in a substantial improvement of 6.6\% in success on zero-shot \texttt{ObjectNav}.

Finally, we qualitatively experiment with an open-world setting and observe that our \texttt{SemanticNav} agents can properly change behavior in response to instructions that include room information. For instance, when finding a \myquote{bathroom sink} the agent does not enter the kitchen, and when looking for a \myquote{kitchen sink} it does not enter bathrooms. Furthermore, we observe similar room awareness patterns for instructions such as \myquote{Find a sink and a stove,} where the target room (\myquote{kitchen}) can be inferred. Source code for reproducing our results will be publicly released.

\section{Related Work}
\label{sec:related-work}

\looseness=-1
Our work builds on research studying image-text alignment techniques (e.g., CLIP~\cite{radford2021clip}) and their use in visual navigation. In this section, we discuss methods most related to our proposed approach.

\paragraph{Image-Text Alignment Models.}

Recent progress in vision-and-language pretraining has led to models such as CLIP~\cite{radford2021clip}, ALIGN~\cite{jia2021align}, and BASIC~\cite{pham2021basic} that can perform open-world image classification, and achieve strong performance on standard computer vision benchmarks (e.g., ImageNet~\cite{deng2009imagenet}). These models learn visual representations by training on massive datasets of image-caption pairs scraped from the web (e.g., the 400M pairs used for CLIP or 6.6B for BASIC). In this work, we take advantage of the semantic representations learned by CLIP to project navigation goals (e.g., a picture of a brown chair or \myquote{brown chair}) into a multimodal, semantic-goal embedding space.

\paragraph{CLIP for Visual Navigation.}

A straightforward approach for using CLIP in a visual navigation agent is to process the agent's observations and navigation instructions (e.g., \myquote{Find a chair}) with the CLIP image and text encoders, then learn a navigation policy that operates on these embeddings. Such a solution was explored in EmbCLIP~\cite{khandelwal2021embclip} with promising results. However, this approach requires ObjectNav rewards or demonstrations to supervise the navigation policy, which is difficult and costly to collect at scale. As a result, existing training datasets tend to be small and agents generalize poorly to new settings. For instance, EmbCLIP only achieves an 8\% success rate in finding objects that were not used in training. By contrast, we train using the image-goal navigation task, which does not require annotated environments. Thus, we are able to scale training to 800 unannotated 3D scenes, which substantially improves generalization (as demonstrated in~\cref{sec:experiments}).

\paragraph{Zero-Shot ObjectNav.}

Two recent works~\cite{al2022zero, gadre2022cow} directly address our motivation (zero-shot \texttt{ObjectNav}) and are most related. First, ZER~\cite{al2022zero} proposes a two-stage framework in which an image-goal navigation (\texttt{ImageNav}) agent is first trained from scratch. Then, independent encoders are trained to map from various modalities (including language) into the image-goal embedding space. A key challenge with this approach is that image-goal embeddings may not capture semantic information because semantic annotations are not used in \texttt{ImageNav} training. Instead, an \texttt{ImageNav} agent trained from scratch may learn to pattern match visual observations and goal image embeddings. By contrast, our approach reverses these two stages, with CLIP pretraining representing stage one. Thus, our approach uses a goal embedding space that captures semantics by design. We empirically demonstrate the benefits of our proposed approach in~\cref{sec:experiments}.

In concurrent work, CLIP-on-Wheels (CoW)~\cite{gadre2022cow} uses a gradient-based visualization technique (GradCAM~\cite{selvaraju2017grad}) with CLIP to localize objects in the agent's observations. This is combined with a heuristic exploration policy to enable zero-shot object-goal navigation. In contrast, we demonstrate that learning a navigation policy can substantially outperform the heuristic exploration approach proposed in~\cite{gadre2022cow} without using explicit object localization techniques.

\section{Preliminaries: Image-Text Alignment and Image-Goal Navigation}

\paragraph{Image-Text Alignment Models.}

Multimodal alignment models aim to learn a mapping from images $v$ and text $t$ into a shared embedding space such that representations for corresponding image-text pairs (e.g., a picture and its caption) are similar. Recent image-text alignment models~\cite{radford2021clip, jia2021align, pham2021basic} use a dual-encoder framework and optimize the InfoNCE~\cite{oord2018representation} contrastive learning objective, which maximizes cosine similarity between representations of matching image-text pairs and minimizes similarity for non-matching pairs. In this work, we leverage CLIP~\cite{radford2021clip}, which was trained on 400M image-text pairs that cover a wide range of visual concepts.

\paragraph{Image-Goal Navigation.}

In image-goal navigation (\texttt{ImageNav})~\cite{zhu2017target}, agents explore an environment to find the position where a goal-image $v^g$ was captured. We consider a setting in which both the goal-image and the agent's observations consist of RGB images taken from the agent's egocentric point of view. An agent can select from four actions: \texttt{MOVE\_FORWARD} by 0.25m, \texttt{TURN\_LEFT} by 30$^{\circ}$, \texttt{TURN\_RIGHT} by 30$^{\circ}$, or \texttt{STOP}. The agent succeeds if it selects \texttt{STOP} within 1.0m of the goal. 

\looseness=-1
An \texttt{ImageNav} episode is uniquely defined by a starting position and (reachable) goal viewpoint within a 3D environment. Thus, \texttt{ImageNav} training data can be procedurally generated without annotating the scene -- i.e., the objects and rooms do not need to be labeled.
As a result, the size of an \texttt{ImageNav} dataset is only limited by the number of environments available for training. In this work, we use \texttt{ImageNav} to train visual navigation agents at scale (in terms of the number of training environments).

\begin{figure}
  \centering
  \includegraphics[width=\textwidth]{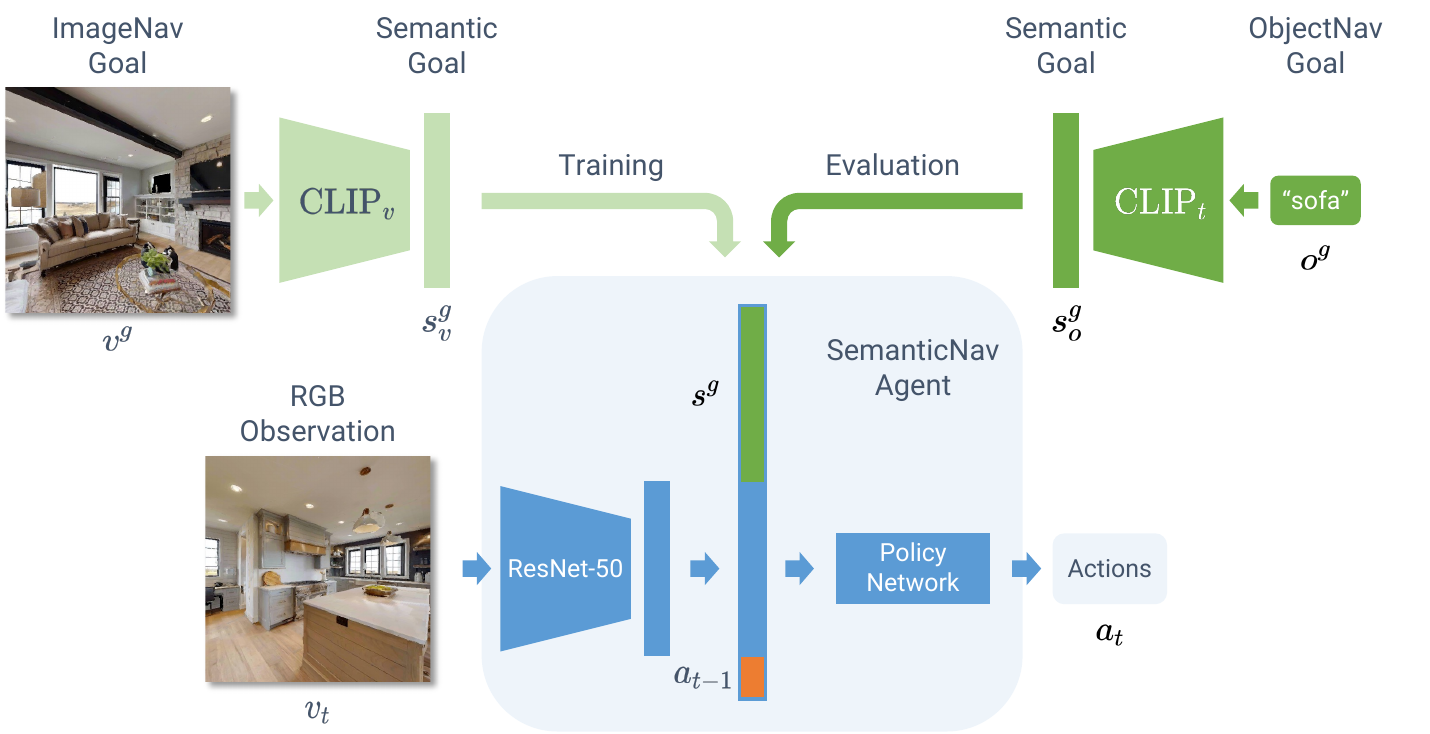}
  \vspace{1ex}
  \caption{We tackle both \texttt{ImageNav} and \texttt{ObjectNav} via a common \texttt{SemanticNav} agent. This agent accepts a semantic goal embedding ($s^g$), which comes from either CLIP's visual encoder ($\texttt{CLIP}_v$) in ImageNav or CLIP's textual encoder ($\texttt{CLIP}_t$) in ObjectNav. Our agent has a simple architecture: RGB observations are encoded with a pretrained ResNet-50, and a recurrent policy network predicts actions using encodings of the goal $s^g$, observation, and the previous action $a_{t-1}$.}
  \label{fig:approach}
\end{figure}

\section{Approach}

This section describes our framework for training visual navigation agents. We use CLIP~\cite{radford2021clip} to produce semantic goal embeddings of image-goals (e.g., a picture of the sink) and object-goals (e.g., \myquote{sink}). This allows training semantic-goal navigation agents at scale using image-goals in HM3D environments~\cite{ramakrishnan2021hm3d}, then deploying these agents for object-goal navigation in a \emph{zero-shot} manner. In other words, our agents execute object-goal navigation without ever directly training for the task.

\subsection{Learning Semantic-Goal Navigation}

As illustrated in~\cref{fig:approach} (top-left), given an image-goal $v^g$, we use a CLIP visual encoder $\texttt{CLIP}_v$ to generate a semantic goal embedding $s_v^g = \texttt{CLIP}_v(v^g)$ that is used to guide navigation. Conceptually, encoding image-goals with CLIP preserves semantic information about the goal, such as visual concepts that might be described in image captions (e.g., \myquote{a sofa in a living room}). However, semantic goal embeddings are less likely to include low-level features (e.g., the exact patterns in a wood floor) that do not correlate with web-scraped captions. While removing low-level information might make the navigation task more difficult, our goal is to learn a policy that transfers to \texttt{ObjectNav} in which agents only receives high-level goals (e.g., \myquote{Find a sofa}). As an added benefit, generating semantic goal embeddings as a pre-processing step substantially improves training time (by $\sim$3.5x).

Our agent architecture is shown in~\cref{fig:approach}. At each timestep $t$, our agent receives an egocentric RGB observation $v_t$ and a goal representation $s_v^g$. The observation is processed by a ResNet-50~\cite{he2016deep} encoder, which is pretrained on the Omnidata Starter Dataset (OSD)~\cite{eftekhar2021omnidata} using self-supervised learning (DINO~\cite{caron2021dino}) following the pretraining recipe presented in OVRL~\cite{yadav2022ovrl}. The output from the ResNet-50 encoder is concatenated with the goal representation $s_v^g$ and an embedding of the agent's previous action $a_{t-1}$ and then passed to the policy network composed of a two-layer LSTM. The policy network outputs a distribution over the action space.

We train our \texttt{SemanticNav} agent with reinforcement learning (RL). During RL training, we use two data augmentation techniques: color jitter and random translation (adapted from~\cite{yadav2022ovrl}). Specifically, we train with DD-PPO~\cite{wijmans2019ddppo} using a reward function proposed for \texttt{ImageNav} by~\citet{al2022zero}:
\begin{equation}
    r_t = r_{\text{success}} + r_{\text{angle-success}} - \Delta_{\text{dtg}} - \Delta_{\text{atg}} + r_{\text{slack}}
\end{equation}
where $r_{\text{success}}=5$ if \texttt{STOP} is called when the agent is within 1m of the goal position (and 0 otherwise), $r_{\text{angle-success}}=5$ if \texttt{STOP} is called when the agent is within 1m of the goal position and the agent is pointing within 25$^{\circ}$ of the goal heading -- i.e., the direction the camera was pointing when the goal image was collected -- (and 0 otherwise), $\Delta_{\text{dtg}}$ is the change in the agent's distance-to-goal -- i.e., the geodesic distance to the goal position, $\Delta_{\text{atg}}$ is the change in the agent's angle-to-goal -- i.e., the difference between the agent's heading and the goal heading -- but is set to 0 if the agent is greater than 1m from the goal, and $r_{\text{slack}}=-0.01$ to encourage efficient navigation. In general, this reward function encourages both reaching the goal and looking towards the goal before calling \texttt{STOP}, which matches the requirements of the downstream \texttt{ObjectNav} task.

\subsection{Zero-Shot Object-Goal Navigation}

Recall that in \texttt{ObjectNav}~\cite{batra2020objectnav}, agents are given a target category (e.g., \myquote{sofa} or \myquote{chair}) and must locate any instance of that object (i.e., \myquote{any sofa} or \myquote{any chair}). Similar to \texttt{ImageNav}, \texttt{ObjectNav} requires exploring new environments that the agent has never seen before. However, in \texttt{ObjectNav}, the goal (e.g., \myquote{sofa}) provides a minimal amount of information about where the agent must go and it requires recognizing any version of the goal object in the new scene.

To address this task, we transform object-goals $o^g$ (e.g., \myquote{sofa}) into semantic goal embeddings using the CLIP text encoder $\texttt{CLIP}_t$, which results in the semantic goal $s_o^g=\texttt{CLIP}_t(o^g)$. CLIP aligns image and text, thus the semantic goals from text $s_o^g$ should be close (in terms of cosine similarity) to the CLIP visual embeddings $s_v^g$ used in training. To keep our approach simple and easily reproducible, we do not use any prompt engineering (e.g., using a template such as ``\texttt{A photo of a <>}''). Instead, we simply use the object name (e.g., \myquote{sofa}) as the object-goal input $o^g$.

\section{Experimental Findings and Qualitative Results}
\label{sec:experiments}

This section studies the zero-shot \texttt{ObjectNav} performance of our proposed approach. First, we evaluate our method in the traditional \texttt{ObjectNav} setting~\cite{batra2020objectnav} where agents must find any instance of the goal object (\myquote{Find a chair}). Then, we explore variations of \texttt{ObjectNav} in which additional information, such as a room location (e.g., \myquote{bathroom sink}), is given to refine the task. These experiments aim to demonstrate both the effectiveness and versatility of our approach.

\subsection{Experimental Setup}

\paragraph{Training Dataset.}

We generate a dataset for training our \texttt{SemanticNav} agent using the 800 training environments from HM3D~\cite{ramakrishnan2021hm3d}. First, we sample 9k \texttt{ImageNav} episodes for each HM3D scan, split equally between 3 difficulty levels corresponding with path length: \textsc{easy} (1.5-3m), \textsc{medium} (3-5m), and \textsc{hard} (5-10m). We follow the episode generation approach from~\cite{mezghani2021memory}. This results in $9\text{k} \times 800 = 7.2\text{M}$ navigation episodes for training. Next, we pre-process the goal-images with the ResNet-50 version of CLIP~\cite{radford2021clip} to produce 1024 dimensional semantic goal vectors $s_v^g$ for each navigation episode. During pre-processing, we further augment the dataset by sampling goal-images at four evenly-spaced heading angles to produce 36M total episodes for training. Sampling at multiple angles approximates the randomized sampling used in~\cite{al2022zero}.

\paragraph{Agent Configurations.}

Two different agent configurations are frequently used in prior work on visual navigation. Configuration \texttt{A} is generally used for \texttt{ImageNav} and has an agent height of 1.5m, radius of 0.1m, and a single 128$\times$128 RGB sensor with a 90$^{\circ}$ horizontal field-of-view (HFOV) placed 1.25m from the ground. Configuration \texttt{B} is typically used for \texttt{ObjectNav} and approximately matches a LoCoBot, with an agent height of 0.88m, radius of 0.18m, and a single 640$\times$480 RGB sensor with a 79$^{\circ}$ HFOV placed 0.88m from the ground. Both configurations use the aforementioned step size of 0.25m and left and right turning angle of 30$^{\circ}$.

\paragraph{Evaluation Datasets.}

We measure performance on one \texttt{ImageNav} and three \texttt{ObjectNav} datasets:
\begin{itemize}[leftmargin=*]
    \setlength\itemsep{0ex}
    \item[--] \texttt{ImageNav} (Gibson) consists of 4,200 episodes from 14 Gibson~\cite{xia2018gibson} validation scenes. The dataset was produced by~\citet{mezghani2021memory} for agents with configuration \texttt{A}.
    \item[--] \texttt{ObjectNav} (Gibson) was generated by~\citet{al2022zero} for agents with configuration \texttt{A}. The dataset consists of 1,000 episodes in 5 Gibson~\cite{xia2018gibson} validation scenes for 6 object categories.
    \item[--] \texttt{ObjectNav} (HM3D), released with the Habitat 2022 challenge, consists of 2,000 episodes from 20 HM3D~\cite{ramakrishnan2021hm3d} validation scenes with objects from 6 categories, and uses agents with configuration \texttt{B}.
    \item[--] \texttt{ObjectNav} (MP3D) released with the Habitat 2020 challenge, contains 2,195 episodes from 11 MP3D~\cite{chang2017mp3d} validation scenes for 21 object categories, and requires agents with configuration \texttt{B}.
\end{itemize}

Due to the different agent configurations required by these evaluation datasets, we train agents with both settings to make fair comparisons with prior work on zero-shot \texttt{ObjectNav}. For all experiments, we report two standard metrics for visual navigation tasks: success rate (\texttt{SR}) and success rate weighted by normalized inverse path length (\texttt{SPL})~\cite{anderson2018evaluation}.

\paragraph{Implementation Details.}

\looseness=-1
We generate a \texttt{SemanticNav} dataset for each agent configuration (\texttt{A} and \texttt{B}). The CLIP ResNet-50 encoder processes 224 $\times$ 224 images. Accordingly, for configuration \texttt{A}, we render 512 $\times$ 512 RGB frames, then resize to 224 $\times$ 224. For configuration \texttt{B}, we render at 640 $\times$ 480, then resize and center crop. We train agents using PyTorch~\cite{pytorch} and the Habitat simulator~\cite{savva2019habitat, szot2021habitat}. Each training run was conducted on a single compute node with 8 NVIDIA A40 GPUs. We train agents for 500M steps, requiring $\sim$1,704 GPU-hours to train two agents (one for each configuration). Additional training hyperparamters are detailed in the Appendix. We report results using the best checkpoint, selected based on \texttt{ObjectNav} validation success rate (\texttt{SR}). During evaluations we sample actions from the agent's output distribution. We report results averaged over three evaluation runs.

\paragraph{Baselines.}

We provide comparisons with the, to the best of our knowledge, only two existing zero-shot methods for object-goal navigation (\texttt{ObjectNav}):
\begin{itemize}[leftmargin=*]
    \setlength\itemsep{0ex}
    \item[--] \textbf{Zero Experience Required (ZER)}~\cite{al2022zero}: first trains an \texttt{ImageNav} agent composed of two ResNet-9 encoders for processing the goal-image and agent observations, and a policy network consisting of a 2-layer GRU. After training the navigation policy, a 2-layer MLP is trained to map from a goal object categories into the goal-image embedding space learned through \texttt{ImageNav} training. This mapping is learned using an in-domain dataset containing 14K images with object category labels.
    \item[--] \textbf{CLIP on Wheels (CoW)}~\cite{gadre2022cow}: builds an occupancy map by projecting depth observations, then searches the environment with frontier-based exploration~\cite{yamauchi1997frontier}. At each step, CoW calculates a 3D saliency map using a depth and RGB observations and the goal object category via Grad-CAM~\cite{selvaraju2017grad}, a gradient-based visualization technique. When the 3D saliency exceeds a threshold the agent navigates to that location and stops. As such, CoW does not require a learned navigation policy.
\end{itemize}

\paragraph{Fully-Supervised ObjectNav.}

To understand the gap to fully-supervised \texttt{ObjectNav} methods, we compare with OVRL~\cite{yadav2022ovrl}, a two-stage framework that achieves state-of-the-art \texttt{ObjectNav} results in our single RGB camera setting. We highlight \textcolor{Periwinkle}{OVRL} in blue to indicate the use of direct supervision.

\begin{table}[t]
\caption{\textbf{Zero-shot ObjectNav performance} on Gibson~\cite{xia2018gibson}, HM3D~\cite{ramakrishnan2021hm3d}, and MP3D~\cite{chang2017mp3d} validation. All methods use a single RGB sensor for agent observations except CoW~\cite{gadre2022cow}, which also uses depth observations and OVRL~\cite{yadav2022ovrl}, which uses \texttt{GPS+Compass} for \texttt{ObjectNav}. Our approach (ZSON) substantially improves on previous zero-shot methods and narrows the gap to SOTA fully-supervised methods such as \textcolor{Periwinkle}{OVRL}~\cite{yadav2022ovrl}, which is not zero-shot and provided for reference. We report ZSON results averaged over three evaluation trials. The standard deviation in ZSON \texttt{ObjectNav} \texttt{SR} is 0.02\% in Gibson, 0.46\% in HM3D, and 0.11\% in MP3D. \textit{$^{*}$indicates reproduced results}}
\label{tab:main-results}
\vspace{2ex}
\begin{subtable}[t]{0.48\textwidth}
\centering
\resizebox{\textwidth}{!}{%
\renewcommand{\arraystretch}{1.3}
\begin{tabular}{c cc cg}
    \toprule
    {} &
    \multicolumn{2}{c}{\makecell{\texttt{ImageNav}\\{\small(Gibson)}}} &
    \multicolumn{2}{c}{\makecell{\texttt{ObjectNav}\\{\small(Gibson)}}} \\
    \cmidrule(l{3pt}r{3pt}){2-3}
    \cmidrule(l{3pt}r{3pt}){4-5}
    Method &
    \texttt{SPL} & \texttt{SR} &
    \texttt{SPL} & \texttt{SR} \\
    \midrule
    \textcolor{Periwinkle}{OVRL}~\cite{yadav2022ovrl} &
    \textcolor{Periwinkle}{27.0\%\phantom{$^{*}$}} & \textcolor{Periwinkle}{54.2\%\phantom{$^{*}$}} & 
    \textcolor{Periwinkle}{-} & \textcolor{Periwinkle}{-} \\
    \midrule
    \phantom{1234} ZER~\cite{al2022zero} \phantom{1234}&
    21.6\%\phantom{$^{*}$} & 29.2\%\phantom{$^{*}$} &
    - & 11.3\% \\
    ZSON (ours) &
    \textbf{28.0\%}\phantom{$^{*}$} & \textbf{36.9\%}\phantom{$^{*}$} &
    12.0\% & \textbf{31.3\%} \\
    \bottomrule
\end{tabular}%
}
\caption{Configuration \texttt{A}}
\label{tab:main-results-a}
\end{subtable}%
\hfill
\begin{subtable}[t]{0.48\textwidth}
\centering
\resizebox{\textwidth}{!}{%
\renewcommand{\arraystretch}{1.3}
\begin{tabular}{c cg cg}
    \toprule
    {} &
    \multicolumn{2}{c}{\makecell{\texttt{ObjectNav}\\{\small(HM3D)}}} &
    \multicolumn{2}{c}{\makecell{\texttt{ObjectNav}\\{\small(MP3D)}}} \\
    \cmidrule(l{3pt}r{3pt}){2-3}
    \cmidrule(l{3pt}r{3pt}){4-5}
    Method &
    \texttt{SPL} & \texttt{SR} &
    \texttt{SPL} & \texttt{SR} \\
    \midrule
    \textcolor{Periwinkle}{OVRL}~\cite{yadav2022ovrl} &
    \textcolor{Periwinkle}{12.3\%$^{*}$} & \textcolor{Periwinkle}{32.8\%$^{*}$} &
    \textcolor{Periwinkle}{\phantom{0}7.0\%} & \textcolor{Periwinkle}{25.3\%} \\
    \midrule
    CoW~\cite{gadre2022cow} (w/depth)&
    - & - &
    \phantom{0}\textbf{6.3\%} & 11.1\% \\
    ZSON (ours) &
    12.6\%\phantom{$^{*}$} & 25.5\%\phantom{$^{*}$} &
    \phantom{0}4.8\% & \textbf{15.3\%} \\
    \bottomrule
\end{tabular}%
}%
\caption{Configuration \texttt{B}}
\label{tab:main-results-b}
\end{subtable}
\end{table}

\subsection{Zero-Shot Object-Goal Navigation}

In~\cref{tab:main-results} we report zero-shot \texttt{ObjectNav} performance. We compare with ZER~\cite{al2022zero} in~\cref{tab:main-results-a} using agent configuration \texttt{A}. Notice that our agent is stronger than ZER on \texttt{ImageNav}, which is the base pretraining task before ObjectNav can be studied. Specifically, we observe a 7.7\% improvement in \texttt{ImageNav} \texttt{SR} (29.2\% $\rightarrow$ 36.9\%). This improvement results from (1) learning to navigate to semantic goal embeddings (as proposed in this work) instead of navigating to image-goal embeddings that are learned from scratch (as done in ZER), (2) using more diverse training environments, and (3) from using a pretrained visual encoder. We provide additional comparisons with ZER using the same set of training environments and without using visual encoder pretraining in~\cref{sec:comparison-zer}, where we also observe improved performance. In~\cref{tab:main-results-a}, we see even larger improvements in \texttt{ObjectNav} \texttt{SR} of 20.0\% (11.3\% $\rightarrow$ 31.3\%). These results indicate that our design decisions are particularly useful for zero-shot \texttt{ObjectNav}.

In~\cref{tab:main-results-b} we compare with CoW~\cite{gadre2022cow} using agent configuration \texttt{B}. In \texttt{ObjectNav} on the MP3D validation set, we find that training a \texttt{SemanticNav} agent improves \texttt{ObjectNav} \texttt{SR} by 4.2\% absolute and 37.8\% relative (11.1\% $\rightarrow$ 15.3\%). These results demonstrate that learning a navigation policy improves zero-shot \texttt{ObjectNav} \texttt{SR} over the hand-designed exploration strategy and stopping criteria proposed by CoW. Moreover, we expect further improvements in zero-shot \texttt{ObjectNav} performance from scaling our approach (e.g., by collecting more training environments). Such scaling is simply not possible with heuristic methods such as CoW because the navigation policy is not learned. The \texttt{SPL} of our approach is 1.5\% lower than CoW. However, unlike CoW, our agent navigates without depth observations, which may reduce path efficiency. On HM3D we find that our agent achieves a strong \texttt{SR} of 25.5\% and \texttt{SPL} of 12.6\%. Impressively, this zero-shot \texttt{SPL} matches \textcolor{Periwinkle}{OVRL}~\cite{yadav2022ovrl}, which is directly trained on 40k human demonstrations~\cite{rramrakhya2022habitatweb} for the \texttt{ObjectNav} task with imitation learning.

\subsection{Comparison with ZER without Encoder Pretraining and Training Environment Diversity}
\label{sec:comparison-zer}

In~\cref{tab:zer}, we train our approach in Gibson environments (instead of HM3D) and do not use a pretrained observation encoder. These settings match ZER~\cite{al2022zero}, allowing for a direct comparison between the two methods. We observe that our approach results in a 4.0\% absolute and 35\% relative improvement in zero-shot \texttt{ObjectNav} success (11.3\% $\rightarrow$ 15.3\%). These results demonstrate that learning to navigate to semantic-goal embeddings outperforms the inverse approach proposed by ZER of first training for image-goal navigation, then learning a mapping from object categories into the image-goal embedding space.

\begin{table}[ht]
\renewcommand{\arraystretch}{1.3}
\caption{\textbf{Comparison with ZER}~\cite{al2022zero} using a ResNet-9 and the Gibson dataset with our approach. Learning \texttt{SemanticNav} (Ours) outperforms learning \texttt{ImageNav} then language grounding (ZER~\cite{al2022zero}).}
\label{tab:zer}
\centering
\vspace{2ex}
\resizebox{0.6\textwidth}{!}{%
\begin{tabular}{c cc cc cg}
    \toprule
    {} & 
    {} & {} &
    \multicolumn{2}{c}{\makecell{\texttt{ImageNav}\\{\small(Gibson)}}} &
    \multicolumn{2}{c}{\makecell{\texttt{ObjectNav}\\{\small(Gibson)}}} \\
    \cmidrule(l{3pt}r{3pt}){4-5}
    \cmidrule(l{3pt}r{3pt}){6-7}
    Method &
    \makecell{Visual\\Encoder} &
    \makecell{Training\\Dataset} &
    \texttt{SPL} & \texttt{SR} &
    \texttt{SPL} & \texttt{SR} \\
    \midrule
    ZER~\cite{al2022zero} & ResNet-9 & Gibson &
    21.6\% & 29.2\% & - & 11.3\% \\
    Ours & ResNet-9 & Gibson &
    \textbf{22.8\%} & \textbf{33.3\%} & \textbf{7.4\%} & \textbf{15.3\%} \\
    \bottomrule
\end{tabular}
}%
\end{table}

\subsection{Additional Ablations}

In~\cref{tab:ablations}, we study the impact of two key design decisions within our method: (1) the visual observation encoder and (2) the number of training environments. While pretraining the visual observation encoder is known to improve visual navigation task performance (demonstrated in~\cite{yadav2022ovrl}), here we study the impacts on zero-shot transfer to \texttt{ObjectNav}. We find that OVRL pretraining improves \texttt{ImageNav} success by 4.5\% (rows 1 vs.\ 3) or 5.8\% (rows 2 vs.\ 4) depending on the dataset used for training. However, the impact on zero-shot \texttt{ObjectNav} performance is substantially larger. Specifically, \texttt{ObjectNav} success improves by 9.4\% (rows 1 vs.\ 3) and 10.4\% (rows 2 vs.\ 4). These results suggest that a strong visual encoder is often essential for zero-shot transfer to \texttt{ObjectNav}.

In rows 3 vs.\ 4, we switch the training dataset from the 72 Gibson~\cite{xia2018gibson} training environments (row 3) to the 800 (unannotated) HM3D~\cite{ramakrishnan2021hm3d} training environments. Surprisingly, we observe a 0.9\% drop in \texttt{ImageNav} success, yet a 6.6\% improvement in \texttt{ObjectNav} success (rows 3 vs.\ 4). A similar trend is observed in rows 1 vs. 2. These trends indicate that training environment diversity is particularly useful for zero-shot \texttt{ObjectNav}.

\begin{table}[ht]
\renewcommand{\arraystretch}{1.3}
\caption{\textbf{Ablations} of the visual encoder and dataset used for training our \texttt{SemanticNav} agents.}
\label{tab:ablations}
\centering
\vspace{2ex}
\resizebox{0.8\textwidth}{!}{%
\begin{tabular}{c cc cc cg}
    \toprule
    {} & {} & {} &
    \multicolumn{2}{c}{\makecell{\texttt{ImageNav}\\{\small(Gibson)}}} &
    \multicolumn{2}{c}{\makecell{\texttt{ObjectNav}\\{\small(Gibson)}}} \\
    \cmidrule(l{3pt}r{3pt}){4-5}
    \cmidrule(l{3pt}r{3pt}){6-7}
    \texttt{\#} &
    \makecell{Visual\\Encoder} &
    \makecell{Training\\Dataset} &
    \texttt{SPL} & \texttt{SR} &
    \texttt{SPL} & \texttt{SR} \\
    \midrule
    \texttt{1} & ResNet-9 from scratch & Gibson &
    22.8\% & 33.3\% & 7.4\% & 15.3\% \\
    \texttt{2} & ResNet-9 from scratch & HM3D &
    23.4\% & 31.1\% & 9.5\% & 20.9\% \\
    \midrule
    \texttt{3} & OVRL (ResNet-50, pretrained) & Gibson &
    27.6\% & \textbf{37.8\%} & 10.0\% & 24.7\% \\
    \texttt{4} & OVRL (ResNet-50, pretrained) & HM3D &
    \textbf{28.0\%} & 36.9\% & \textbf{12.0\%} & \textbf{31.3\%} \\
    \bottomrule
\end{tabular}
}%
\end{table}

\begin{figure}[ht]
  \centering
  \includegraphics[width=\textwidth]{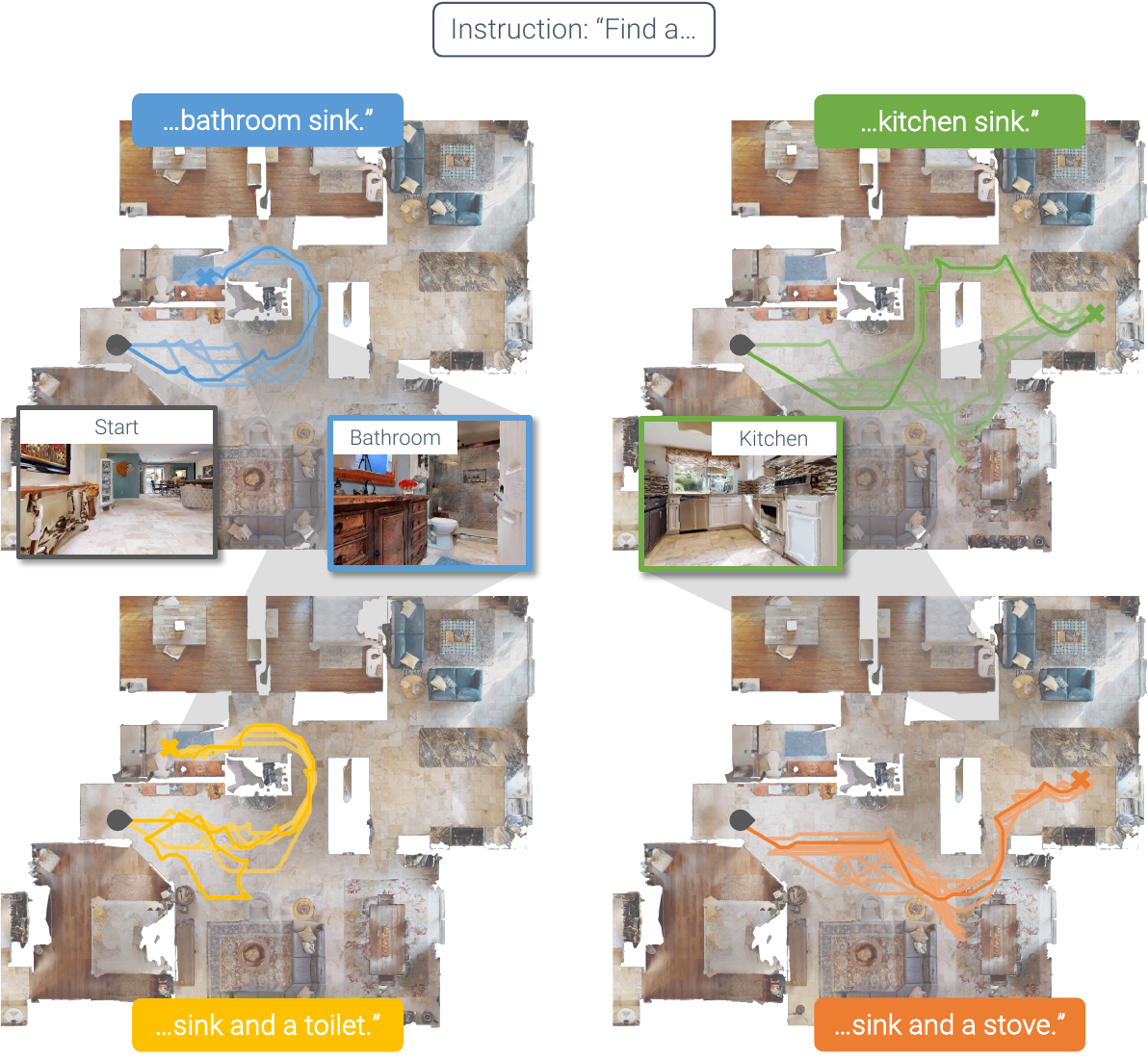}
  \vspace{1ex}
  \caption{\textbf{Qualitative examples} for navigating to complex object descriptions. For each trail, the agent is spawned at the \emph{start} position looking into the house (i.e., to the right on the maps) and given one of four instructions. Each instruction is run five times with the path for the first trail highlighted in bold colors. Our agent appropriately navigates to the correct rooms, demonstrating an understanding of both explicit (\textcolor{mygreen}{\myquote{Find a kitchen sink}}) and implicit (\textcolor{myorange}{\myquote{Find a sink and a stove}}) room information.}
  \label{fig:qualitative}
\end{figure}

\subsection{Qualitative Analysis}

\looseness=-1
In~\cref{fig:qualitative}, we present qualitative examples of our agent navigating to more complex object descriptions (e.g., \textcolor{myblue}{\myquote{Find a bathroom sink}}). In each trial, the agent starts at the same position and heading (next to the front door looking into the house). The only thing that changes about the initial conditions is the instructions given to the agent (\myquote{Find a...} \textcolor{myblue}{\myquote{...bathroom sink}}, \textcolor{mygreen}{\myquote{...kitchen sink}}, \textcolor{myyellow}{\myquote{...sink and a toilet}}, or \textcolor{myorange}{\myquote{...sink and a stove}}). Since the agent's policy is stochastic, we show 5 sampled rollouts and highlight the first run in bold colors.

We find that given room information such as \textcolor{myblue}{\myquote{bathroom}} or \textcolor{mygreen}{\myquote{kitchen}}, the agent appropriately finds a \myquote{sink} in the corresponding rooms in the house. Furthermore, in these examples the agent does not enter the \textcolor{mygreen}{\myquote{kitchen}} when prompted to look for a \textcolor{myblue}{\myquote{bathroom sink,}} and vice-versa. In these long trajectories (ranging from 88 to 225 steps), we observe more exploration in the living room and direct navigation when target rooms are visible. We qualitatively observe interesting learned behaviors -- for instance, the agent often performs a 360$^{\circ}$ turn before navigating, possibly to survey the environment.

Next, we experiment with variations in which room information can be inferred from the instruction, but is not explicit. We use \textcolor{myyellow}{\myquote{sink and a toilet}} to indicate \textcolor{myblue}{\myquote{bathroom}} and \textcolor{myorange}{\myquote{sink and a stove}} for \textcolor{mygreen}{\myquote{kitchen}}. In these examples, we discover that our agent still navigates to the correct rooms, suggesting that it learns some priors of indoor spaces, such as that a \textcolor{myorange}{\myquote{stove}} is often found within a \textcolor{mygreen}{\myquote{kitchen.}}

\section{Discussion}

\looseness=-1
We present a \emph{zero-shot} method for learning \emph{open-world} object-goal navigation (\texttt{ObjectNav}). Our approach involves projecting image-goals into a semantic-goal embedding space using an image-and-text alignment model (CLIP). This creates a semantic-goal navigation task that does not require annotated 3D environments or collecting human demonstrations. Thus, our method is easy to scale. We discover that \texttt{SemanticNav} agents outperform previous zero-shot \texttt{ObjectNav} methods, and we identify two factors that have a strong impact on navigation success -- pretraining the visual encoder and training in a diverse set of environments. In an open-world setting, we observe navigation patterns that suggest that \texttt{SemanticNav} agents can understand complex instructions, such as \myquote{Find a sink and a stove.}

\paragraph{Limitations and Impact.}

\texttt{SemanticNav} agents appear to learn useful priors of indoor environments such as which room contains a \myquote{stove.} However, agents may struggle in scenes where a navigation target is in an unusual location (e.g., a stove in a bedroom). Biases in the 3D environments used to train such agents might exaggerate these issues and affect deployments in non-traditional settings. Thus, interventions to mitigate this problem should be considered. Future work might explore how to use the natural language interface to \texttt{SemanticNav} agents to guide exploration in such scenarios.

\section*{Acknowledgements}

The Georgia Tech effort was supported in part by NSF, ONR YIP, ARO PECASE, and ARL. The views and conclusions contained herein are those of the authors and should not be interpreted as necessarily representing the official policies or endorsements, either expressed or implied, of the U.S. Government, or any sponsor.

{
\small
\bibliographystyle{unsrtnat}
\bibliography{references}
}

\clearpage
\section*{Checklist}

\begin{enumerate}

\item For all authors...
\begin{enumerate}
  \item Do the main claims made in the abstract and introduction accurately reflect the paper's contributions and scope?
    \answerYes{}
  \item Did you describe the limitations of your work?
    \answerYes{}
  \item Did you discuss any potential negative societal impacts of your work?
    \answerYes{}
  \item Have you read the ethics review guidelines and ensured that your paper conforms to them?
    \answerYes{}
\end{enumerate}

\item If you are including theoretical results...
\begin{enumerate}
  \item Did you state the full set of assumptions of all theoretical results?
    \answerNA{}
        \item Did you include complete proofs of all theoretical results?
    \answerNA{}
\end{enumerate}

\item If you ran experiments...
\begin{enumerate}
  \item Did you include the code, data, and instructions needed to reproduce the main experimental results (either in the supplemental material or as a URL)?
    \answerYes{}
  \item Did you specify all the training details (e.g., data splits, hyperparameters, how they were chosen)?
    \answerYes{}
        \item Did you report error bars (e.g., with respect to the random seed after running experiments multiple times)?
    \answerYes{}
        \item Did you include the total amount of compute and the type of resources used (e.g., type of GPUs, internal cluster, or cloud provider)?
    \answerYes{}
\end{enumerate}

\item If you are using existing assets (e.g., code, data, models) or curating/releasing new assets...
\begin{enumerate}
  \item If your work uses existing assets, did you cite the creators?
    \answerYes{}
  \item Did you mention the license of the assets?
    \answerYes{}
  \item Did you include any new assets either in the supplemental material or as a URL?
    \answerYes{}
  \item Did you discuss whether and how consent was obtained from people whose data you're using/curating?
    \answerYes{}
  \item Did you discuss whether the data you are using/curating contains personally identifiable information or offensive content?
    \answerYes{}
\end{enumerate}

\item If you used crowdsourcing or conducted research with human subjects...
\begin{enumerate}
  \item Did you include the full text of instructions given to participants and screenshots, if applicable?
    \answerNA{}
  \item Did you describe any potential participant risks, with links to Institutional Review Board (IRB) approvals, if applicable?
    \answerNA{}
  \item Did you include the estimated hourly wage paid to participants and the total amount spent on participant compensation?
    \answerNA{}
\end{enumerate}

\end{enumerate}

\appendix

\begin{table}[ht]
\renewcommand{\arraystretch}{1.2}
\caption{\textbf{Comparison of ObjectNav methods.} Open-world methods are not limited to a closed set of object categories. Zero-shot methods do not use \texttt{ObjectNav} annotations for training.}
\label{tab:comparisons}
\centering
\vspace{1ex}
\begin{tabular}{ccc}
\toprule
Method & Open-World & Zero-Shot \\
\midrule
\makecell{Fully-Supervised Methods~\cite{rramrakhya2022habitatweb, chaplot2020object, ye2021auxiliary, maksymets2021thda, yadav2022ovrl}} & \textcolor{Red}{\xmark} & \textcolor{Red}{\xmark} \\
EmbCLIP~\cite{khandelwal2021embclip} & \textcolor{Green}{\cmark} & \textcolor{Red}{\xmark} \\
ZER~\cite{al2022zero} & \textcolor{Red}{\xmark} & \textcolor{Green}{\cmark} \\
CoW~\cite{gadre2022cow} & \textcolor{Green}{\cmark} & \textcolor{Green}{\cmark} \\
\rowcolor{Gray}
ZSON (ours) & \textcolor{Green}{\cmark} & \textcolor{Green}{\cmark} \\
\bottomrule
\end{tabular}
\end{table}

\section{Extended Discussion of Related Work}

\looseness=-1
In ~\cref{tab:comparisons}, we compare object-goal navigation (\texttt{ObjectNav}) methods along two dimensions: \emph{open-world} and \emph{zero-shot}. Open-world methods are not restricted to object categories from a closed predetermined vocabulary (e.g., \myquote{chair}, \myquote{bed}, \myquote{sofa}, etc.). Zero-shot methods do not use \texttt{ObjectNav} annotations (e.g., labeled environments or large-scale human demonstrations~\cite{rramrakhya2022habitatweb}) for training. 

\looseness=-1
Traditional, fully-supervised \texttt{ObjectNav} methods~\cite{rramrakhya2022habitatweb, chaplot2020object, ye2021auxiliary, maksymets2021thda, yadav2022ovrl} rely on a closed-world assumption and task-specific training data -- i.e., they are neither open-world nor zero-shot. EmbCLIP~\cite{khandelwal2021embclip} can be used for open-world \texttt{ObjectNav} because it pre-processes object-goals (e.g., \myquote{chair}) with a CLIP~\cite{radford2021clip} text encoder $\text{CLIP}_t$. Thus, the EmbCLIP interface allows describing objects using the open-vocabulary supported by $\text{CLIP}_t$. However, EmbCLIP is trained directly with \texttt{ObjectNav} annotations based on labeled 3D environments. Consequently, EmbCLIP is not a zero-shot method.

\looseness=-1
The method proposed in~\cite{al2022zero} (ZER) is zero-shot because it uses the image-goal navigation (\texttt{ImageNav}) task for training. However, ZER cannot perform open-world \texttt{ObjectNav} because it relies on a mapping from a closed-set of object categories into the image-goal embedding space for transfer to object-goal navigation. CoW~\cite{gadre2022cow} is a zero-shot method that does not require training a navigation policy. Instead, CoW uses a heuristically defined policy that has no ability to learn about indoor layouts of home environments (e.g., the fact that \myquote{stoves} are found in \myquote{kitchens} as illustrated in~\cref{fig:qualitative}). However, CoW is able to perform open-world \texttt{ObjectNav} through the use of CLIP visual and text encoders.

By contrast, our approach (ZSON) uses CLIP to project image-goals and object-goals into a common semantic-goal embedding space, which converts \texttt{ImageNav} and \texttt{ObjectNav} into semantic-goal navigation (\texttt{SemanticNav}). This enables training with semantic-goals derived from images, followed by \emph{zero-shot} transfer to \emph{open-world} \texttt{ObjectNav}.

An interesting direction for future work might be to train \texttt{SemanticNav} agents with multi-task training using semantic-goals derived from both image- and object-goals. Such a solution would not be zero-shot. However, it might combined the advantages of large-scale training with image-goals (as used in our approach) with the advantages of smaller-scale task-specific training with object-goals (as used in EmbCLIP). We present initial results in this direction in~\cref{sec:finetune}.

\begin{table}[ht]
\renewcommand{\arraystretch}{1.3}
\caption{\textbf{Results of finetuning} with ObjectNav annotations. \textit{$^{*}$indicates reproduced results}}
\label{tab:finetune-results}
\centering
\vspace{1ex}
\resizebox{0.55\textwidth}{!}{%
\begin{tabular}{c cc cg}
    \toprule
    \texttt{\#} &
    Method &
    Dataset &
    \texttt{SPL} & \texttt{SR} \\
    \midrule
    \texttt{1} & OVRL~\cite{yadav2022ovrl} & MP3D &
    7.0\% & \textbf{25.3\%}\\ 
    \texttt{2} & ZSON (ours) & MP3D &
    4.8\% & 15.3\% \\
    \texttt{3} & ZSON finetuned 25M steps (ours) & MP3D &
    \textbf{9.2\%} & 22.9\% \\
    \midrule
    \texttt{4} & OVRL~\cite{yadav2022ovrl} & HM3D &
    12.3\%$^{*}$ & 32.8\%$^{*}$\\
    \texttt{5} & ZSON (ours) & HM3D &
    12.6\%\phantom{$^{*}$} & 25.5\%\phantom{$^{*}$} \\
    \texttt{6} & ZSON finetuned 100M steps (ours) & HM3D &
    \textbf{27.0\%}\phantom{$^{*}$} & \textbf{49.6\%}\phantom{$^{*}$} \\
    \bottomrule
\end{tabular}
}%
\end{table}

\section{Results of ObjectNav Finetuning}
\label{sec:finetune}

In this section, we study the benefits of additional task-specific training by finetuning ZSON agents using \texttt{ObjectNav} annotations -- i.e., manually labeled training environments. Specifically, we initialize with a \texttt{SemanticNav} agent trained for 500M steps of experience in HM3D environments using image-goals. Then, we finetune for ObjectNav in either MP3D~\cite{chang2017mp3d} or HM3D~\cite{ramakrishnan2021hm3d} annotated environments using reinforcement learning (RL) with the finetuning approach from~\cite{ilrl_baselines}. 

In~\cref{tab:finetune-results}, we find that finetuning ZSON agents results in 7.6\% - 24.1\% absolute improvements in \texttt{ObjectNav} success rates (SR). Specifically, in row 3 we finetune for 25M steps in MP3D environments. This leads to a 7.6\% absolute improvement in SR (15.3\% $\rightarrow$ 22.9\%) and a 4.4\% absolute improvement in SPL (4.8\% $\rightarrow$ 9.2\%). This 9.2\% SPL surpasses the state-of-the-art in the RGB-only setting of 7.0\% SPL, which was set by OVRL~\cite{yadav2022ovrl} (row 1) using direct supervision from 40k human demonstrations in MP3D environments. Similarly, in row 6, we finetune for 100M steps in HM3D environments. This results in a 24.1\% absolute improvement in SR (25.5\% $\rightarrow$ 49.6\%) and a 14.4\% absolute improvement in SPL (12.6\% $\rightarrow$ 27.0\%). These results exceed the OVRL~\cite{yadav2022ovrl} baseline presented in row 4 (which was trained in MP3D environments and is identical to the agent in row 1) by 16.8\% absolute in SR (32.8\% vs. 49.6\%) and 14.7\% absolute in SPL (12.3\% vs. 27.0\%).

\begin{table}[ht]
\renewcommand{\arraystretch}{1.3}
\caption{\textbf{Additional ablations} of the visual encoders used for training our \texttt{SemanticNav} agents.}
\label{tab:additional-ablations}
\centering
\vspace{1ex}
\resizebox{0.6\textwidth}{!}{%
\begin{tabular}{c cc cc cg}
    \toprule
    {} & {} & {} &
    \multicolumn{2}{c}{\makecell{\texttt{ImageNav}\\{\small(Gibson)}}} &
    \multicolumn{2}{c}{\makecell{\texttt{ObjectNav}\\{\small(Gibson)}}} \\
    \cmidrule(l{3pt}r{3pt}){4-5}
    \cmidrule(l{3pt}r{3pt}){6-7}
    \texttt{\#} &
    Encoder &
    Dataset &
    \texttt{SPL} & \texttt{SR} &
    \texttt{SPL} & \texttt{SR} \\
    \midrule
    \texttt{1} & ResNet-9 (scratch) & HM3D &
    23.4\% & 31.1\% & 9.5\% & 20.9\% \\
    \texttt{2} & ResNet-50 (scratch) & HM3D &
    22.4\% & 28.2\% & 6.9\% & 16.6\% \\
    \midrule
    \texttt{3} & ResNet-50 (OVRL) & HM3D &
    \textbf{28.0\%} & \textbf{36.9\%} & \textbf{12.0\%} & \textbf{31.3\%} \\
    \bottomrule
\end{tabular}
}%
\end{table}

\section{Additional Ablation Experiments}

\looseness=-1
\cref{tab:ablations} contains ablations of the visual encoder used to process RGB observations within our agent. In~\cref{tab:additional-ablations}, we provide additional ablations that compare the ResNet-9 and ResNet-50 architectures with and without pretraining -- i.e., from scratch vs. using OVRL~\cite{yadav2022ovrl}. We find that without pretraining, switching to the larger ResNet-50 encoder leads to a 2.9\% drop in \texttt{ImageNav} \texttt{SR} and a 4.3\% drop in \texttt{ObjectNav} \texttt{SR} (rows 1 vs. 2). By contrast, using OVRL pretraining, performance improves on all tasks across all metrics. For instance, \texttt{ImageNav} \texttt{SR} improves by 5.8\% and \texttt{ObjectNav} \texttt{SR} improves by 10.4\% (rows 1 vs. 3). 

\begin{table}[ht]
\renewcommand{\arraystretch}{1.1}
\caption{\textbf{Hyperparameters} used to train \texttt{SemanticNav} agents.}
\label{tab:hyperparameters}
\centering
\vspace{1ex}
\begin{tabular}{lc}
\toprule
Parameter & Value \\
\midrule
Number of GPUs & 8 \\
Number of environments per GPU & 32 \\
Rollout length & 64 \\
PPO epochs & 2 \\
Number of mini-batches per epoch & 2 \\
Optimizer & Adam \\
\quad{}Learning rate & $2.25\times10^{-4}$ \\
\quad{}Weight decay & $1.0\times10^{-6}$ \\
\quad{}Epsilon & $1.0\times10^{-5}$ \\
PPO clip & 0.2 \\
Generalized advantage estimation & True \\
\quad{}$\gamma$ & 0.99 \\
\quad{}$\tau$ & 0.95 \\
Value loss coefficient & 0.5 \\
Max gradient norm & 0.2 \\
DDPPO sync fraction & 0.6 \\
\bottomrule
\end{tabular}
\end{table}

\section{Additional Training Details}

\cref{tab:hyperparameters} details the default hyperparameters used in all of our training runs. For ablation experiments in Gibson~\cite{xia2018gibson} environments we reduce the ``number of environments per GPU'' to 28 for ResNet-9 experiments and 20 for ResNet-50 due to GPU memory constraints. We use the ResNet-9 implementation from~\cite{al2022zero} and ResNet-50 implementation from~\cite{savva2019habitat}.

\section{Additional Qualitative Results}

\begin{figure}[t]
  \centering
  \includegraphics[height=0.42\textheight]{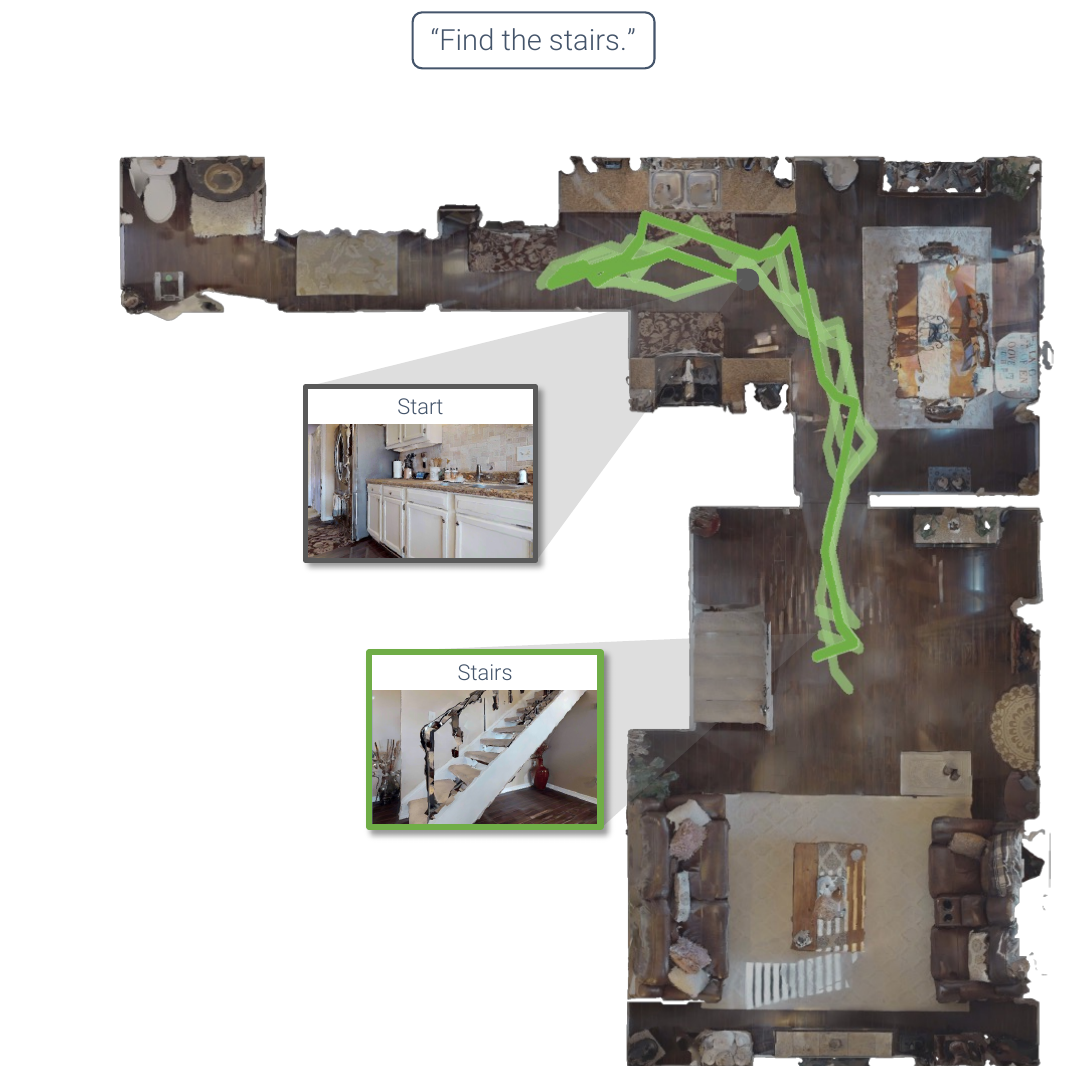}
  \vspace{1ex}
  \caption{Qualitative example of successful navigation to the \myquote{stairs.} The number of steps taken by our agent over five trials ranges from 80 to 102.}
  \label{fig:appendix-stairs}
\end{figure}

\begin{figure}[ht]
  \centering
  \includegraphics[width=\textwidth]{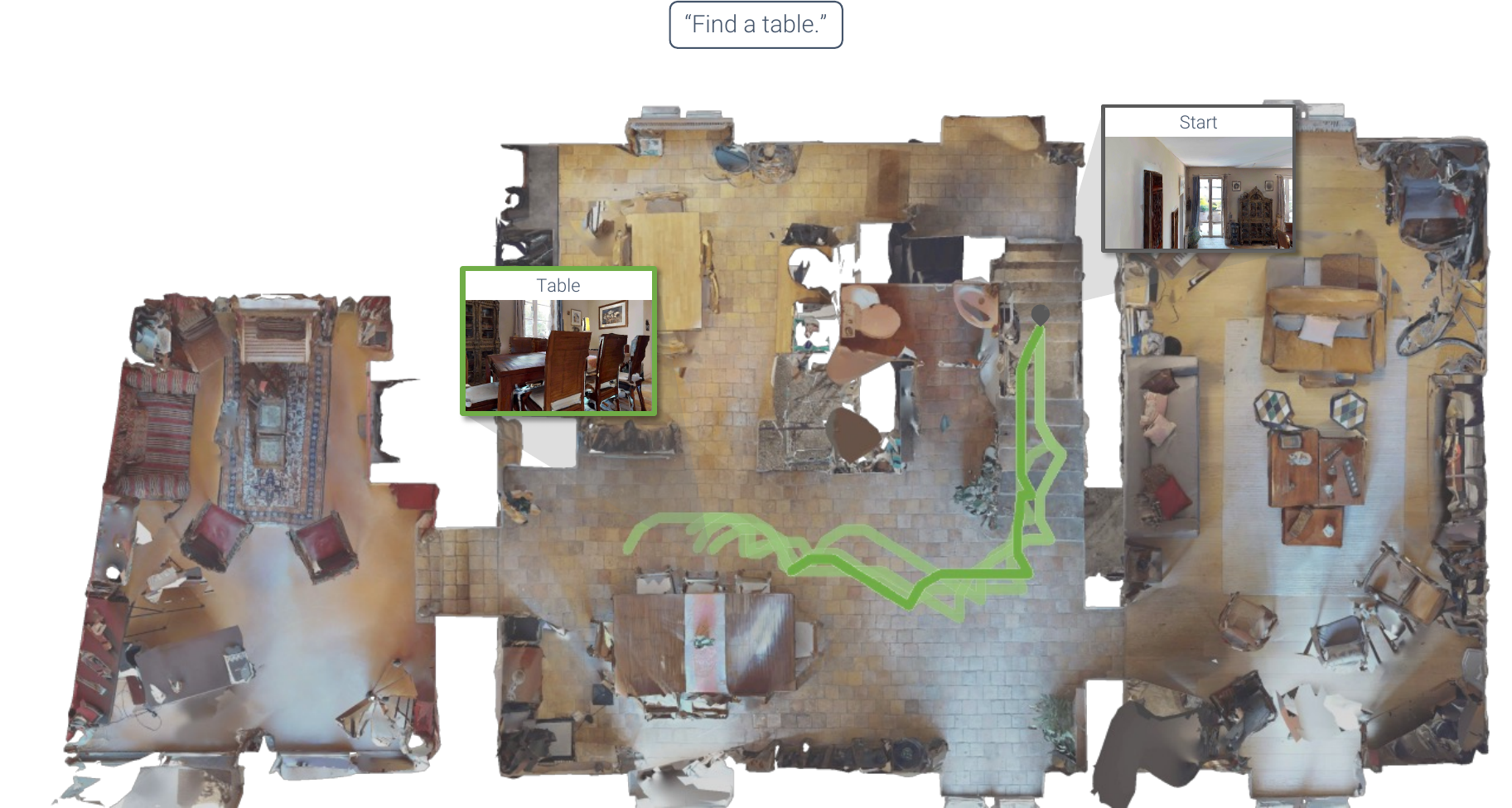}
  \vspace{1ex}
  \caption{Qualitative example of successful navigation to a \myquote{table.} The number of steps taken by our agent over five trials ranges from 60 to 98.}
  \label{fig:appendix-table}
\end{figure}

\begin{figure}[ht]
  \centering
  \includegraphics[height=0.42\textheight]{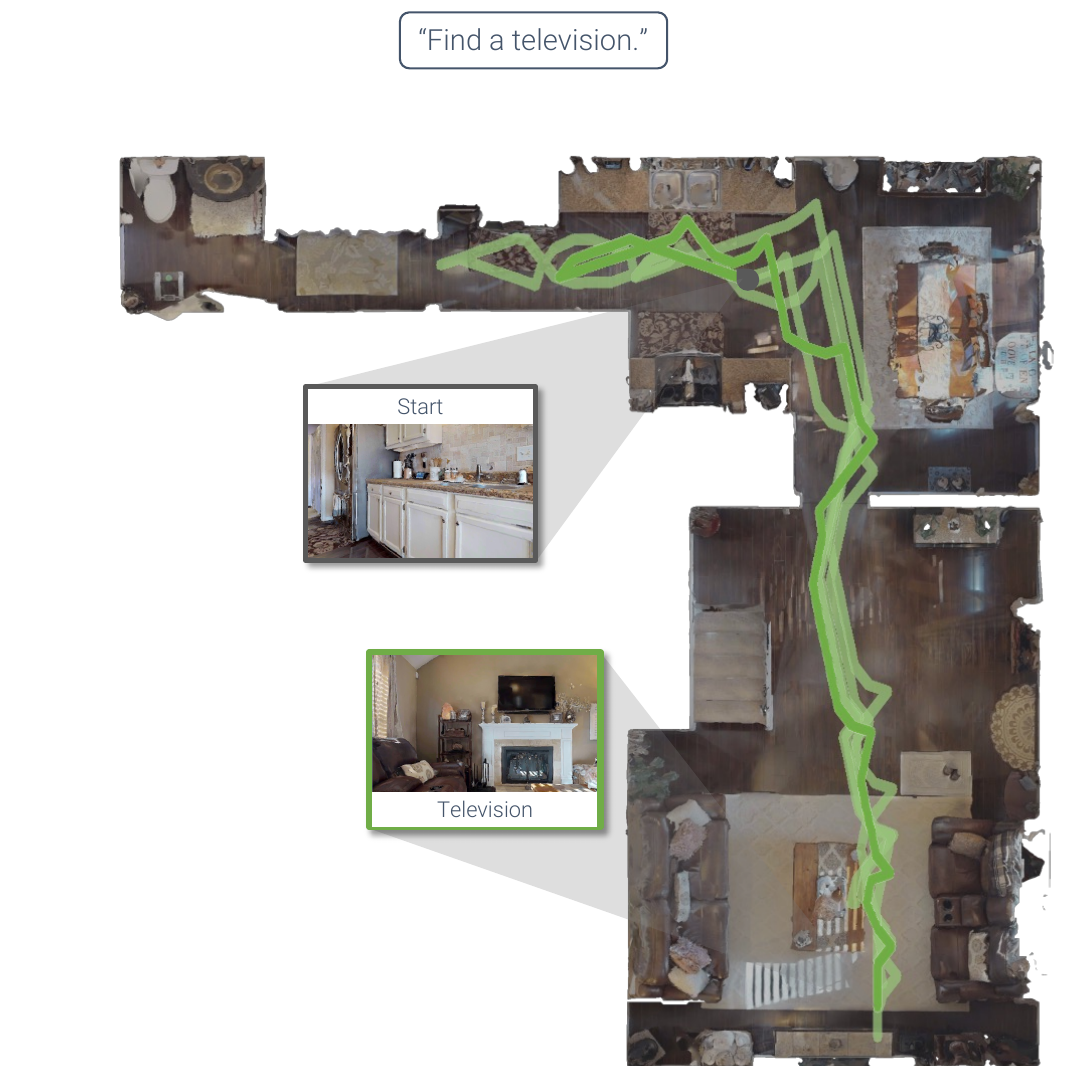}
  \vspace{1ex}
  \caption{Qualitative example of successful navigation to a \myquote{television.} The number of steps taken by our agent over five trials ranges from 78 to 148.}
  \label{fig:appendix-television}
\end{figure}

\begin{figure}[ht]
  \centering
  \includegraphics[height=0.42\textheight]{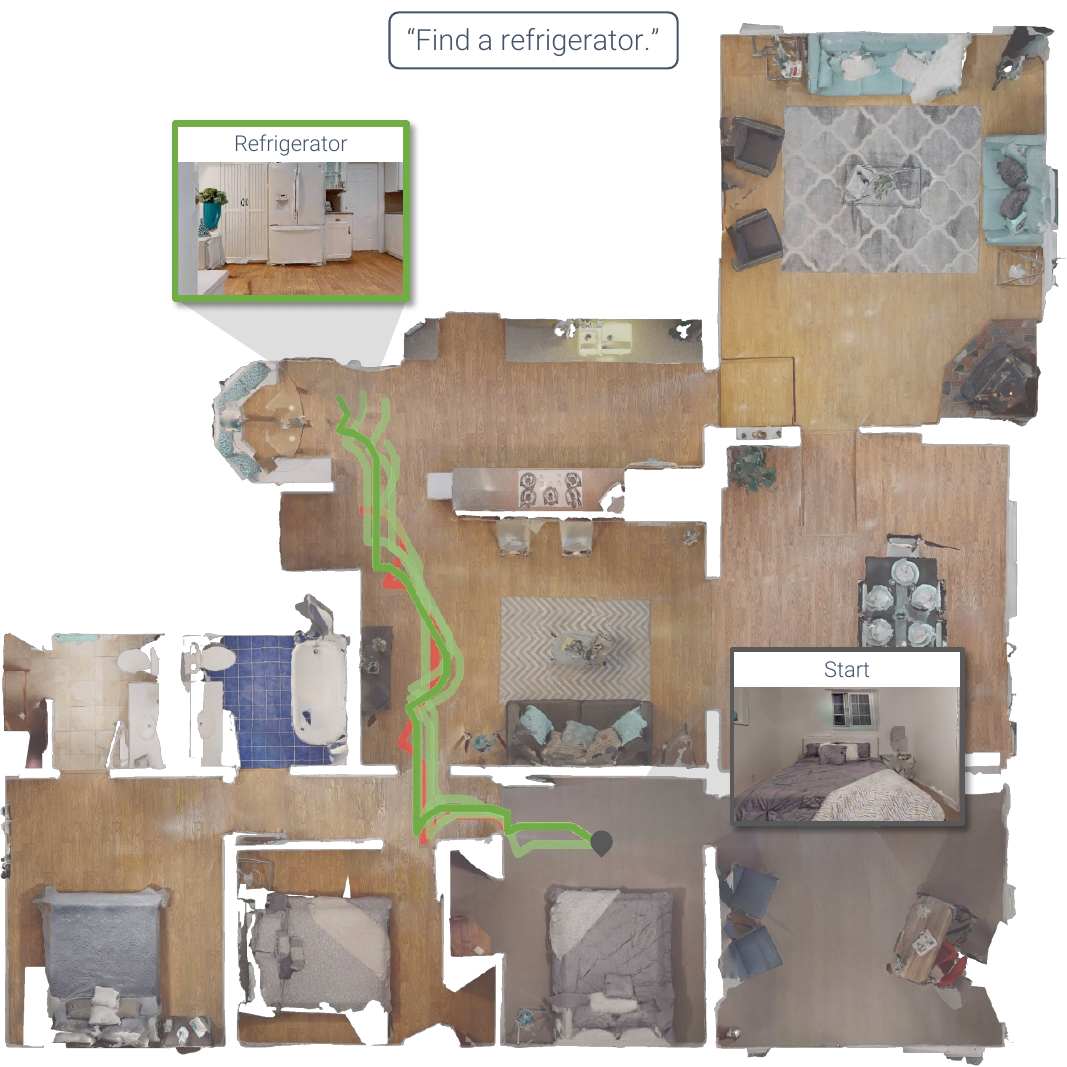}
  \vspace{1ex}
  \caption{Qualitative example of \myquote{Find a refrigerator.} The agent succeeds in 4 of 5 trials (\textcolor{mygreen}{green}) from the same starting position. In the failure (\textcolor{red}{red}) the agent stop short of fridge. The number of steps ranges from 83 to 99.}
  \label{fig:appendix-refrigerator}
\end{figure}

\begin{figure}[ht]
  \centering
  \includegraphics[height=0.42\textheight]{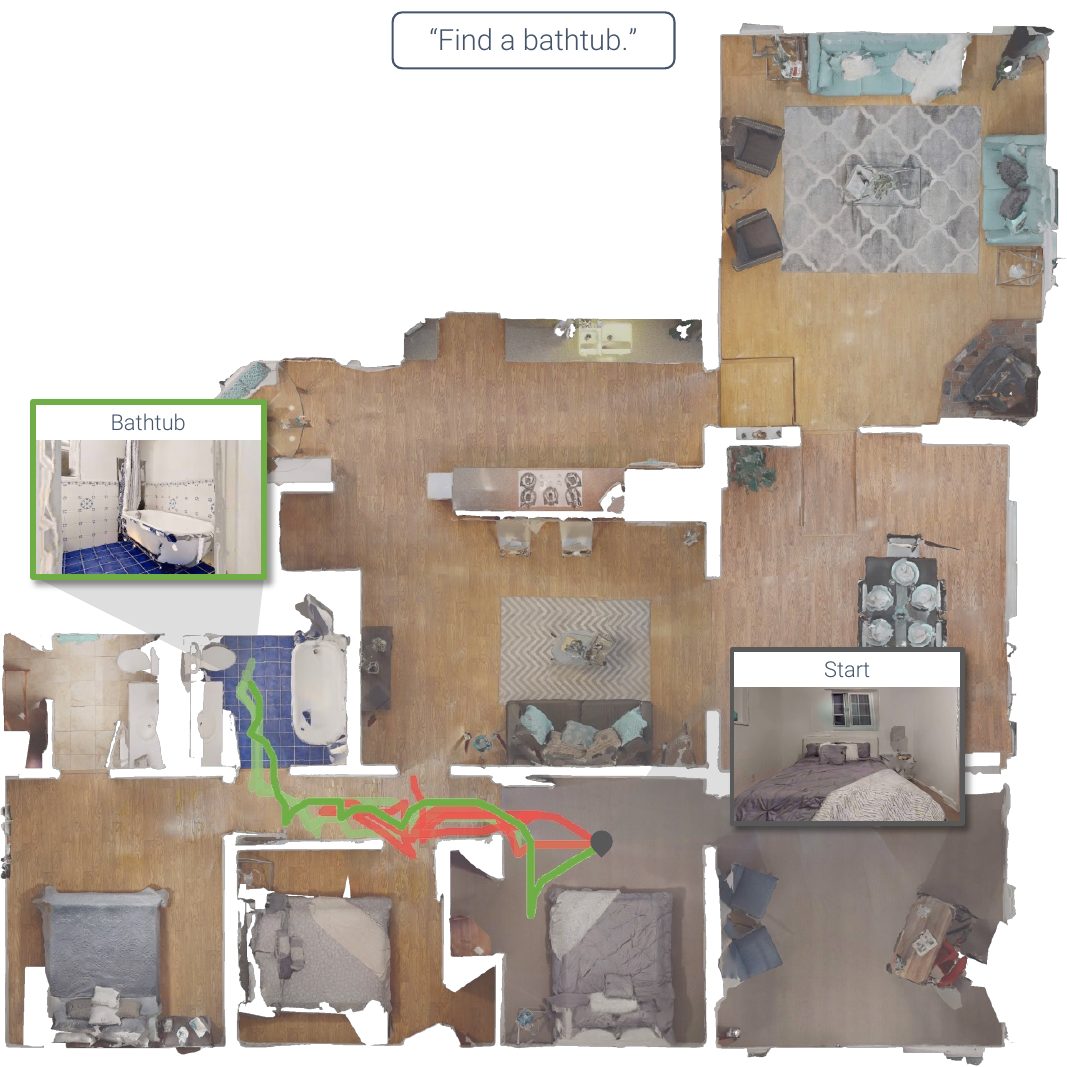}
  \vspace{1ex}
  \caption{Qualitative example of \myquote{Find a bathtub.} The agent succeeds in 3 of 5 trials (\textcolor{mygreen}{green}) from the same starting position. In the two failures (\textcolor{red}{red}) the agent never enters the bathroom. The number of steps ranges from 58 to 114.}
  \label{fig:appendix-bathtub}
\end{figure}

\begin{figure}[ht]
  \centering
  \includegraphics[width=\textwidth]{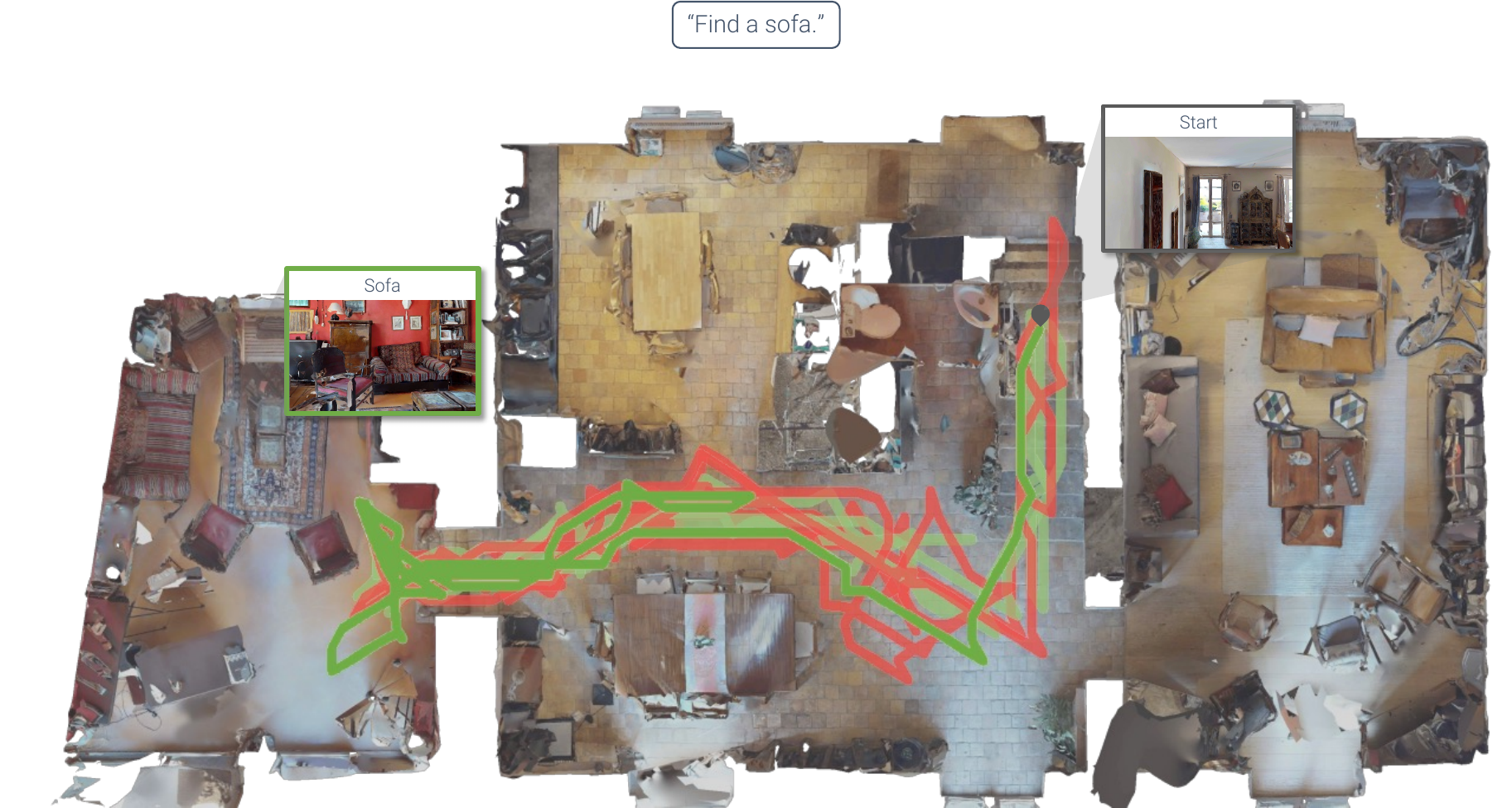}
  \vspace{1ex}
  \caption{Qualitative example of \myquote{Find a sofa.} The agent succeeds in 3 of 5 trials (\textcolor{mygreen}{green}) from the same starting position. In the two failures (\textcolor{red}{red}) the agent stops in the dining area (center). The agent never enters the room to the right with two \myquote{sofas}. The number of steps ranges from 174 to 501.}
  \label{fig:appendix-sofa}
\end{figure}

\begin{figure}[ht]
  \centering
  \includegraphics[height=0.42\textheight]{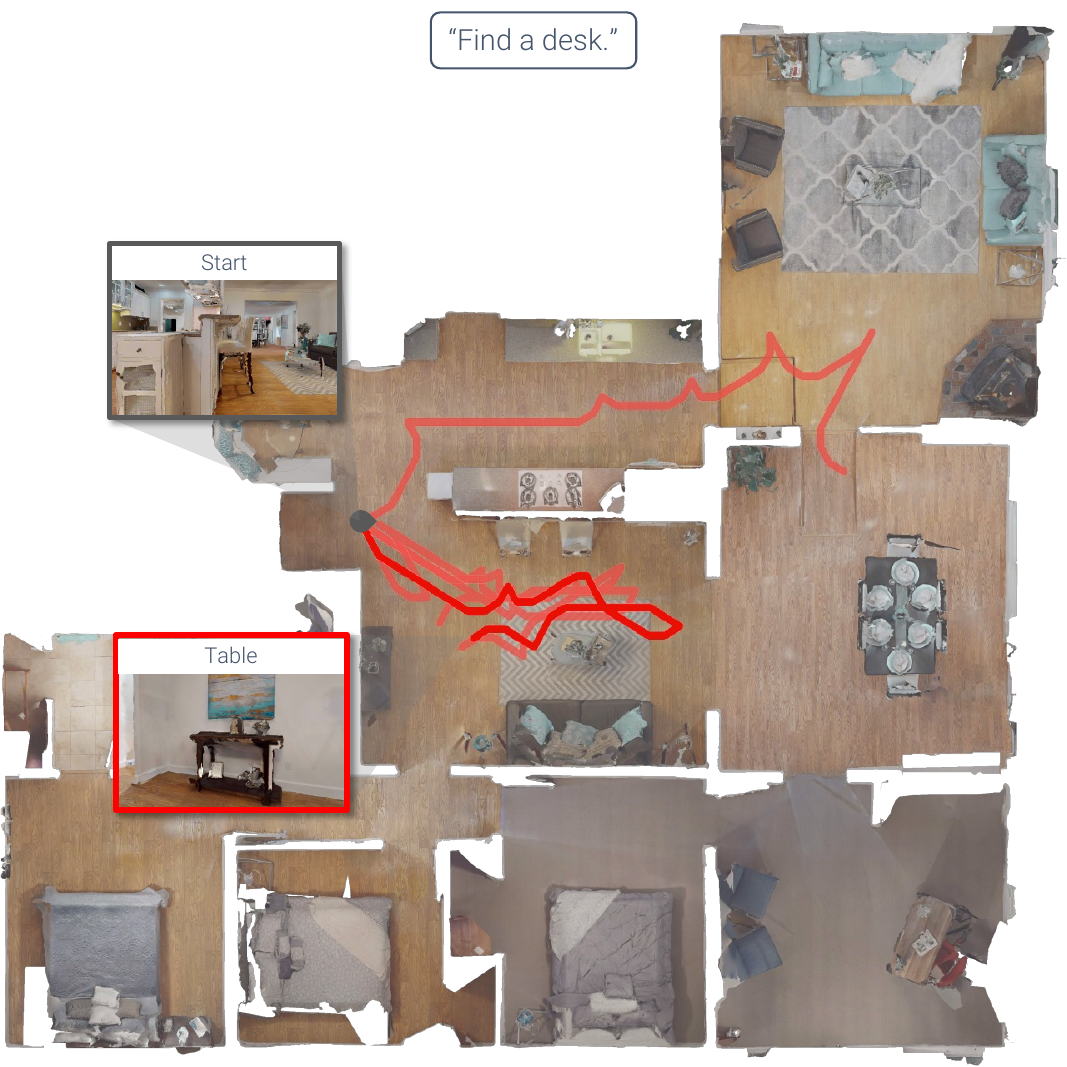}
  \vspace{1ex}
  \caption{Qualitative example of \myquote{Find a desk.} The agent fails in all five trials (\textcolor{red}{red}), stopping at a table in 3 of 5 runs. The agent never enters the room in the bottom right that contains a \myquote{desk}. The number of steps ranges from 49 to 114.}
  \label{fig:appendix-desk}
\end{figure}

\begin{figure}[ht]
  \centering
  \includegraphics[width=\textwidth]{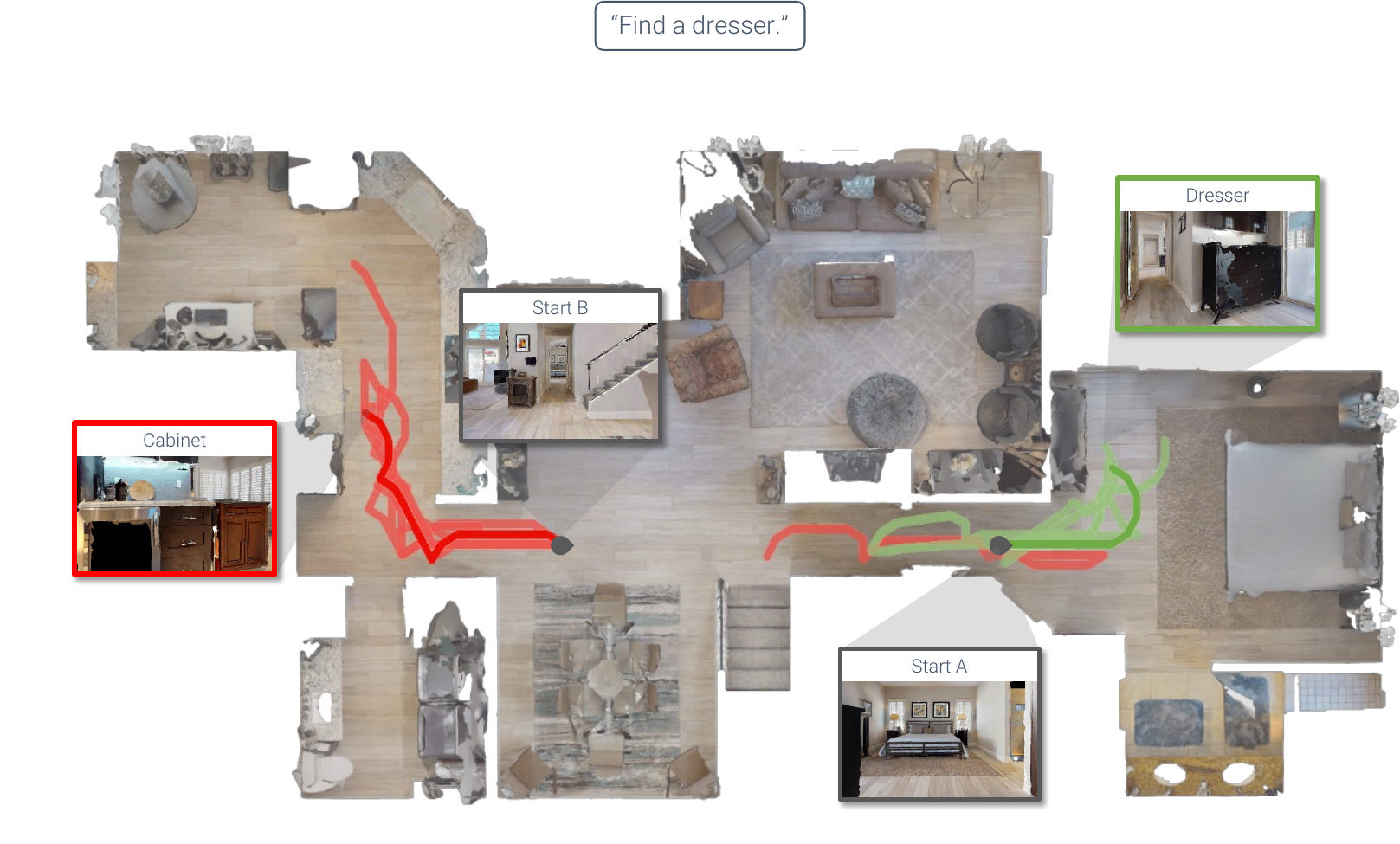}
  \vspace{1ex}
  \caption{Qualitative example of \myquote{Find a dresser.} We run five trails from two starting positions (A and B). From Start A the agent is able to find a dresser in 4 of 5 trials (\textcolor{mygreen}{green}). However, when the starting location is shifted further from the bedroom (Start B) the agent enters the kitchen and fails in all five runs (\textcolor{red}{red}), stopping near the kitchen cabinets. These failures might be due to the similarity in appearance between cabinets and dressers. The number of steps ranges from 29 to 109.}
  \label{fig:appendix-dresser}
\end{figure}

In~\crefrange{fig:appendix-stairs}{fig:appendix-dresser}, we provide additional qualitative results. \crefrange{fig:appendix-stairs}{fig:appendix-television} show successful navigation to \myquote{stairs}, \myquote{table}, and \myquote{television} (respectively), highlighting the versatility of our ZSON agent.

\crefrange{fig:appendix-refrigerator}{fig:appendix-dresser} illustrate various failure modes of the learned policy. In~\cref{fig:appendix-refrigerator}, the agent successfully finds a \myquote{refrigerator} in 4 out of 5 trials. However, in one case the agent stops before entering the kitchen, despite having a view of the \myquote{refrigerator}. In this case, the failure mode is stopping short.

In~\cref{fig:appendix-bathtub}, the agent successfully finds a \myquote{bathtub} when it enters the bathroom nearest to the starting position. However, in 2 of 5 trials it fails to find that bathroom, thus does not find a \myquote{bathtub.} In this case, the failures are due to poor exploration. Similarly, in~\cref{fig:appendix-sofa} the agent does find a seating area that resembles a \myquote{sofa} in 3 of 5 trials. However, it stops in the dining area in 2 trials. Furthermore, it never enters the room to the right, which contains two sofas -- again, demonstrating poor exploration.

In~\cref{fig:appendix-desk} the agent never enters the room to the bottom right, which contains a \myquote{desk,} thus failing in all five runs. In 3 of 5 trials it stops near a table that does not appear to be functioning as a desk (e.g., there is no chair nearby). In these examples, the agent might be confusing objects that can have a similar appearance. Finally, in~\cref{fig:appendix-dresser}, we start the agent from two different positions (A and B) and provide the instruction \myquote{Find a dresser.} When the agent is initialized near the bedroom (A) it is able to find the dresser in 4 out of 5 trials. However, from position B the agent navigates into the kitchen and stops near the cabinets. Again, these failures may be due to similarities in the appearance of dressers and cabinets.

\end{document}